\documentclass[nohyperref]{article}

\PassOptionsToPackage{}{natbib}

\usepackage{microtype}
\usepackage{graphicx}
\usepackage{subfigure}
\usepackage{booktabs} 
\usepackage{multirow}

\usepackage{overpic}
\usepackage{pict2e}

\usepackage[preprint]{neurips_2023}

\usepackage[utf8]{inputenc} 
\usepackage[T1]{fontenc}    
\usepackage{hyperref}

\usepackage{amsfonts}       
\usepackage{nicefrac}       
\usepackage{xcolor}         

\usepackage{amsmath}
\usepackage{algorithm}
\usepackage[noend]{algpseudocode}
\usepackage{scalerel,stackengine}
\stackMath
\definecolor{new}{HTML}{DD0000}
\definecolor{old}{HTML}{7D7D7D}
\newcommand{\mname}{Swift-DynGFN }

\usepackage{tikz}

\title{Causal Inference in Gene Regulatory Networks \\ with GFlowNet: Towards Scalability in Large Systems}

\usepackage{authblk}

\author[$\ $ 1,2]{\textbf{Trang Nguyen} \thanks{Work done under an internship at Mila Institute and National University of Singapore. Correspondence to $\mathsf{trang.nguyen@mila.quebec}$. }}
\author[1,3]{\textbf{Alexander Tong}}
\author[1,3]{\textbf{Kanika Madan}}
\author[$\ $ 1,3,4]{\authorcr\textbf{Yoshua Bengio}\thanks{Equal contribution.}}
\author[$\dagger$1,2]{\textbf{Dianbo Liu}}
\affil[1]{Mila -- Quebec AI Institute}
\affil[2]{National University of Singapore}
\affil[3]{Université de Montréal}
\affil[4]{CIFAR AI Chair}

\begin{document}

\maketitle

\begin{abstract}

Understanding causal relationships within Gene Regulatory Networks (GRNs) is essential for unraveling the gene interactions in cellular processes. However, causal discovery in GRNs is a challenging problem for multiple reasons including the existence of cyclic feedback loops and uncertainty that yields diverse possible causal structures. Previous works in this area either ignore cyclic dynamics (assume acyclic structure) or struggle with scalability. We introduce \mname as a novel framework that enhances causal structure learning in GRNs while addressing scalability concerns. Specifically, \mname exploits gene-wise independence to boost parallelization and to lower computational cost. Experiments on real single-cell RNA velocity and synthetic GRN datasets showcase the advancement in learning causal structure in GRNs and scalability in larger systems. 
\end{abstract}




\section{Introduction}

Gene regulatory networks (GRNs) play a critical role in cell biology, orchestrating a highly intricate interplay of molecular interactions that ultimately govern the behavior of cells \citep{grn1, grns}. Understanding causality entails deciphering the intricate web of relationships between genes, shedding light on how the activity of one gene can intricately influence the expression or behavior of another \citep{causal_in_grn2, causal_in_grn3, causal_in_grn, 10.1093/bib/bbad281, article_1}. However, uncovering causality within GRNs faces challenges, including the cyclic feedback loops  \citep{feedback_loop, atanackovic2023dyngfn} and uncertainty that leads to multiple possible causal structures \citep{atanackovic2023dyngfn, uncertainty1, uncertainty}. Prior efforts in this domain have encountered two restrictions. Firstly, most previous works formulate the problem as a Directed Acyclic Graph (DAG) \citep{article_2, lorch2021dibs, deleu2022bayesian, DBLP:journals/corr/abs-2106-07635}, thus overlooking the cyclic nature inherent in these networks. Secondly, scalability remains a concern when dealing with large GRNs \citep{atanackovic2023dyngfn, deleu2022bayesian}.   

The \textit{Bayesian dynamic structure learning} framework introduced in \cite{atanackovic2023dyngfn} and the incorporation with Generative Flow Networks (GFlowNets) \citep{bengio2021flow, gfn} sheds light on solving the first limitation. Particularly, to model causality within GRNs, \citealp{atanackovic2023dyngfn} simplifies structure learning in GRNs as a sparse identification problem within a dynamical system, utilizing RNA velocity data to estimate the rate of change of a gene's expression. 
In the context of dynamical systems, it is feasible to depict both the causal relations between variables and the system's changing behavior through time. 
Besides, GFlowNet plays a critical role in the work to model intricate distributions over cyclic structures.
However, the scalability is still one restriction of their work on larger systems. 

In this study, we propose \mname to continue to navigate the intricacies of causal structures in GRNs and address scalability. \mname improves the architecture of GFlowNet and draws upon the \textit{Bayesian dynamic structure learning} framework. Notably, we enhance causal structure learning by optimizing variable-wise influence, leveraging predictions from previous variables, and allowing the model to decide on the variable order to be processed.
Furthermore, we improve the scalability by parallelizing the prediction process, reducing the trajectory length from $n^2$ to $n$ (where $n$ is the number of variables), significantly reducing the time and space requirements. 

The main contributions of this work are summarized as 
(1) Introduce \mname that improves causal structure learning in GRNs and tackles the scalability challenge, 
(2) Showcase the capability to capture uncertainty and accurately infer causal relationships within GRNs on real single-cell velocity, and
(3) Prove the scalability potential of Swift-DynGFN on the synthetic GRN data.

\section{Preliminaries}
\subsection{Dynamical systems in Single-cell RNA Velocity}
Denote a finite dataset as $\mathcal{D}$, containing dynamic pairs $(x, dx)$, where $x$  represents a state in a time-invariant stochastic dynamical system and $dx$ denotes its time derivative. 
In estimating the change rate in gene expression by leveraging \textit{RNA velocity} \cite{velocitymethod}, $x$ and $dx$ align to \textit{gene expression levels} and \textit{velocity} of changes in gene expression, respectively.  
We aim to learn posterior over explanatory graphs, $G$, that defines the sparsity graph pattern among variables in $\mathcal{D}$, represented as $Q(G|\mathcal{D})$. 

\subsection{Generative Flow Networks} \label{main:gfn} Generative Flow Networks (GFlowNet) constitute a probabilistic framework designed to facilitate the generation of a diverse array of candidates over spaces of discrete objects \citep{deleu2022bayesian, bengio2021flow, gfn}. Particularly, GFlowNet learns a probabilistic strategy for building structured objects, \textit{e.g.} a graph, by generating a sequence of actions that gradually transform a partial object to a complete object by taking a sequence of actions where each action corresponds to adding an edge. Several training objectives have been used \citep{deleu2022bayesian, bengio2021flow, malkin2022trajectory, Madan2022LearningGF}, one of which is the \textit{detailed balance} objective \cite{gfn}. 
\begin{equation} \label{eq:flow_matching}
              F(s)P_F( s'|s) = F(s')P_B(s|s') 
\end{equation}

\textbf{Detailed Balance (DB) Training Objective} \quad Denoting a GFlowNet model parameterized by $\psi$ that optimizes \textit{forward policy} $P_F(s'|s, \psi)$ and \textit{backward policy} $P_B(s'|s, \psi)$ corresponding to \textit{Markovian flow} of a non-terminal state $F_{\psi}(s)$, the \textit{DB constraint} \cite{gfn} is presented as Equation \ref{eq:flow_matching} in transformation $s \rightarrow s'$. Subsequently, the \textit{detail balance loss} is presented in Equation \ref{eq:main_loss} that optimizes the \textit{DB constraint}. With a terminal state $s_n$, a \textit{Reward matching loss} \cite{gfn, malkin2022trajectory} is added, formulated as
$\mathcal{L}_R(s_n) = (\log(R(s_n)) - \log(F_\psi(s_n)))^2$ with $R(s_n)$ implies to the reward obtained at state $s_n$. 
\begin{equation} \label{eq:main_loss}  
    \mathcal{L}_{DB}(s_{i-1},  s_i) = \bigg( \log \frac{F_{\psi}(s)P_F(s_i|s_{i-1} , \psi)}{F_{\psi}(s_i)P_B(s_{i-1} |s_i, \psi)} \bigg)^2
\end{equation}

\section{Proposed method: \mname}
Intuitively, \mname enhances the variable-wise influence in predicting causal structures, raising parallelization for each variable and conducting sequential computing among different variables to alleviate time and space requirements. 

Facilitating causal influence, we tackle two questions: 
(1) "\textbf{What has been done?}", means the prediction of the current variable is affected by predictions made on other nodes that ensures variables causally contribute to others and 
(2) "\textbf{What's next?}", means \mname selects the next node to be processed that harnesses the causal relationships in the precision of predictions for each other.

\begin{figure}[t]
\centering
\includegraphics[angle=-90, width=0.99\textwidth] {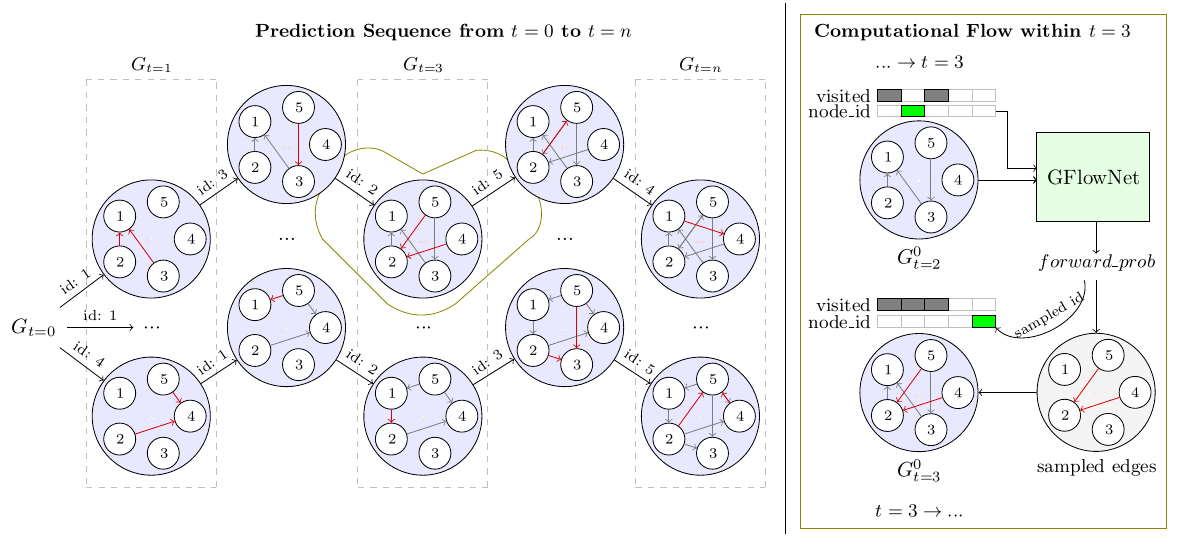}
\caption{\textbf{\mname Intuition and Computation Flow}. Edges added from the previous turns are gray, and newly added edges are red. The model acknowledges \textit{all} previously added edges and outputs (1) \textit{all} incoming edges of the current node and (2) the node for the next turn ($\mathrm{node\_id}$). The $\mathrm{visited}$ binary mask marks nodes are done. A batch of graphs is predicted in parallel.} 
\label{fig:tikz_example}
\end{figure}
In terms of computational complexities, \mname design two strategies:
(1) the prediction of incoming edges to the current node of interest is parallel, drastically reducing processing time and
(2) we cut off the computation decomposition into each variable as observed in prior work \citep{atanackovic2023dyngfn} to avoid parameter outbreak when scaling up the number of nodes. 

\subsection{Variable-wise causal influence in \mname}
Regarding paying attention to predictions made on previous variables, we formulate $Q(G|D)$ as in Equation \ref{eq:qgd}. Particularly, $G \in     \mathbb{R}^{B \times  (n+1) \times (n \times n)}$ denotes states of $B$ causal structures during $t\in [0\dots n]$ steps, which have a shape of $n\times n$ per graph, where $n$ is the number of variables. From the implementation point of view, the GFlowNet model takes incoming edges on visited nodes and the current node's index inputs. 
\begin{equation} \label{eq:qgd}
Q(G|D) = \prod_{i \in 0\dots n} Q(G_{t=i}|G_{t=i-1}, D)   
\end{equation}
To decide the next variable to be processed, a variable's index that has not been visited yet is sampled as an action by the forward policy, along with the incoming edges of the current variable of interest. We employ the forward probability to sample both actions, with a binary mask to mask the set of visited nodes when selecting the next variable. In the first turn ($t=0$), only the action of the next variable index is taken into account, whereas incoming edges are not sampled since the node of interest has not been selected yet.

\subsection{Parallelization in \mname} 
Different from prior approaches \cite{atanackovic2023dyngfn, deleu2022bayesian, bengio2021flow, malkin2022trajectory, Madan2022LearningGF} that sample a single edge at a time, we sample all incoming edges of the current node from the same forward probability. Subsequently, the time complexity is dramatically optimized from $n^2$ to $n$ regarding the length of the prediction trajectory.
In addition, we avoid designing a GFlowNet in each node, as it introduces a potential threat to parameter explosion in a larger number of nodes.
Instead, we operate a single GFlowNet model shared among variables, reducing the number of parameters required in larger systems.

\subsection{Optimization strategy in \mname}
We utilize the \textit{DB objective} as presented in Section \ref{main:gfn} with the terminal state defined as $G_n$ and
reward as $R(G_d) = e^{-||dx_b - \widehat{dx_b}||^2_2 + \lambda_0||G_n||_0}$
where $dx_b$ is the ground truth of the input batch $x_b$, and the $\lambda_0||G_n||_0$ term encourage sparsity of the GRNs.

\section{Experiments}
We conduct experiments to verify two hypotheses:  $\mathcal{H}_1$ - Leveraging the causal relationship among genes enhances causal inference in GRN  (Section \ref{main:grn_res}) and $\mathcal{H}_2$- \mname reduces time and space complexities, contributing to improvements in large-scale systems (Section \ref{main:synth_res}).

We compare our method to DynGFN, DynBCD, and DynDiBS, reported in DynGFN paper \cite{atanackovic2023dyngfn}. Particularly, besides GFlowNet, \citealp{atanackovic2023dyngfn} integrates BCD \citep{bcd} and DiBS \citep{lorch2021dibs}, which are designed for static systems,  to \textit{Bayesian dynamic structure learning}, denoted as DynBCD and DynDiBS. In addition, we employ Bayes-SHD and AUC metrics to evaluate the predicted structures over the true \textit{graphs}.


\subsection{Experiment on Single-Cell RNA-velocity Data} \label{main:grn_res}

\textbf{Dataset} \quad We investigate the cell cycle dataset of human Fibroblasts  \citep{dataset} that contains records of 5000 cells and more than 10,000 genes. Following \citealp{atanackovic2023dyngfn}, we utilize a group of five genes, where Cdc25A activates Cdk1, which, in turn, inhibits Cdc25C, while the Mcm complex is correlated with Cdc25A but does not directly interact with Cdk1 in the cell cycle regulation.
With this setting, the GRN system contains 81 admissible causal structures.

Supporting $\mathcal{H}_1$, Table \ref{tab:gene} presents the performance and computation details of \mname alongside the reproduction of all baselines, experimented on RNA velocity on a single NVIDIA A100 and 4 CPUs. Overall, our proposed method delivers precise predictions while reducing computational complexities. 
Regarding prediction quality, \mname outperforms all baselines by a notable margin in both metrics, evident by the $0.14$ increased AUC and $0.44$ reduced Bayes-SHD compared to DynGFN. 
Regarding computational complexities, our best configuration utilizes far fewer parameters and GPU hours for training than DynGFN. 
Even when configured to match DynGFN's parameter count, our method maintains superior performance and faster training time.  

\begin{table}[t]
\caption{\textbf{Dynamic causal structure inference in GRN}. Reported scores are mean and std over five seeds. \mname outperforms all baselines and reduces computational complexities.}
\label{tab:gene} 
\begin{center}
\begin{tabular}{l|rr |  cc} 
\hline
 & \multicolumn{4}{c}{\bf Cellular System - RNA Velocity}  \\ 
 \textbf{Methods} & \textbf{Bayes-SHD$\downarrow$} & \textbf{AUC$\uparrow$}  & \textbf{\#Params} & \textbf{Duration (h)} \\ \hline
 DynBCD & \textbf{2.79}$\pm$0.34 & 0.53$\pm$0.07 &  100 & 3.76\\
 DynDiBS & 6.82$\pm$0.78 & 0.46$\pm$0.03 &  50.0k & 1.02\\ 
 DynGFN & 3.37$\pm$0.51 & 0.59$\pm$0.03 &  255.4k & 1.16 \\  
 \mname$_{best} $ & \textbf{2.93}$\pm$0.21 & \textbf{0.73}$\pm$0.04 &  87.4k & 0.53 \\ 
 \mname$_{large} $  &    3.22$\pm$0.36 &  0.69$\pm$0.03 &  255.4k & 0.99 \\
\hline
\end{tabular}
\end{center}
\end{table}

\begin{figure}[t]
  \centering
  \includegraphics[height=0.219\textwidth]{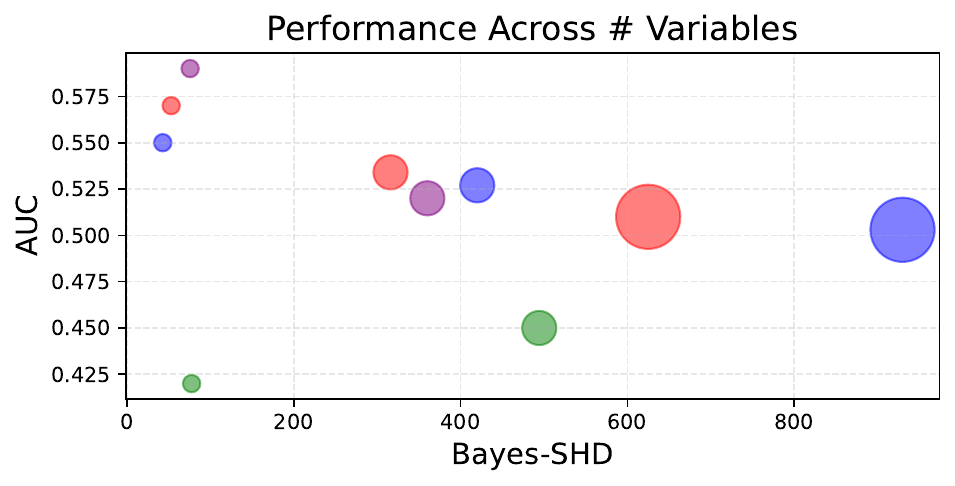}  
  \includegraphics[height=0.219\textwidth]{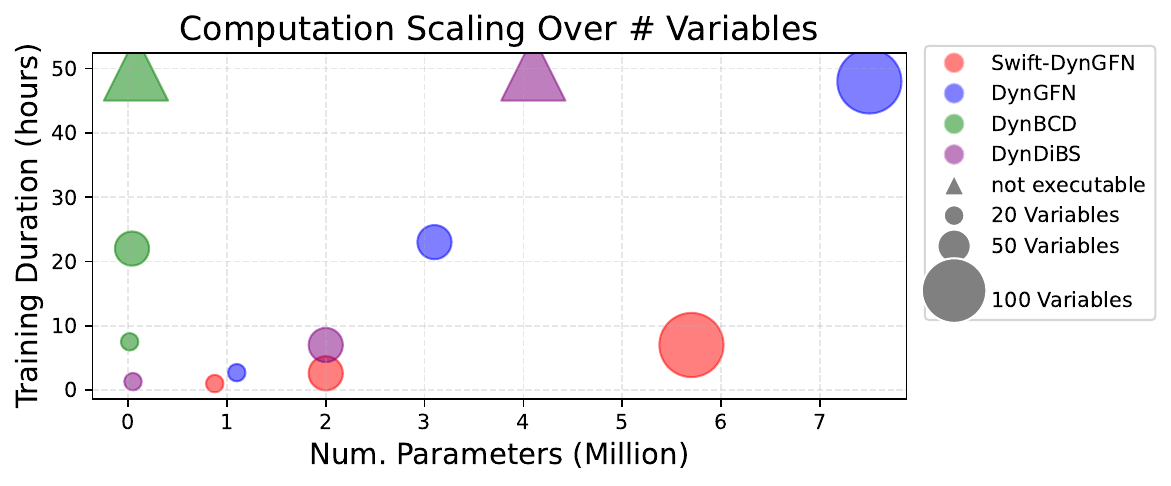} 
  \caption{\textbf{Scalability comparison on synthetic data.} \mname shows strong scalability as it avoids complexities outbreak in large-scale systems while producing remarkable performance. } 
  \label{fig:scalability}
\end{figure} 

\subsection{Experiment on Synthetic Data} \label{main:synth_res} 

\textbf{Dataset} \quad  We adopt the dataset generation using the indeterminacy model from \citealp{atanackovic2023dyngfn} (more detail is in Appendix \ref{ap:indeterminacy_model}) and create a non-linear dynamical system $dx = \mathsf{sigmoid}(\mathbf{A}x)$. We design the number of variables to vary from $20$, $50$, and $100$ with a fixed sparsity equal to $0.9$. 

Examining $\mathcal{H}_2$, Figure \ref{fig:scalability} illustrates the \mname scalability, conducted on 2 NVIDIA A100-80GB GPUs and 8 CPUs. \mname consistently demonstrates robust scalability, requiring fewer computational resources in large-scale systems while delivering considerable performance.
On the left, \mname outperforms or is on par with baselines as the number of nodes increases. 
On the right, \mname avoids computational outbreak, a phenomenon observed in baselines as the system scales up. For instance, with 100 nodes, DynGFN's training duration is five times longer than that of Swift-DynGFN, while DynBCD and DynDiBS struggle with CUDA memory requirements.

\section{Conclusion} 
In this study, we proposed \mname that improves dynamical causal structure inference in large-scale systems. We leverage the GFlowNet model to strengthen the causal consideration while simultaneously parallelizing the computation to reduce time and space requirements. Our experiments on GRN and synthetic data illustrate \mname's effectiveness in improving causal inference ability and scalability in dealing with large-scale systems compared to baselines. 

\textbf{Future Works}: Potential directions to further improve this study include (1) extending the number of genes to estimate ability in large-scale GRN systems and (2) investigating the causal structure in diverse biological contexts, such as Metabolic Pathways and Immune Response. 

\section*{Acknowledgement}
We gratefully acknowledge the support received for this research. This research was enabled in part by computational resources provided by Mila. Each member involved in this research is funded by their primary institution. 

\bibliography{nips_ws}

\begin{thebibliography}{31}
\providecommand{\natexlab}[1]{#1}
\providecommand{\url}[1]{\texttt{#1}}
\expandafter\ifx\csname urlstyle\endcsname\relax
  \providecommand{\doi}[1]{doi: #1}\else
  \providecommand{\doi}{doi: \begingroup \urlstyle{rm}\Url}\fi

\bibitem[Ahmed et~al.(2020)Ahmed, Roy, and Kalita]{causal_in_grn3}
Syed Ahmed, Swarup Roy, and Jugal Kalita.
\newblock Assessing the effectiveness of causality inference methods for gene regulatory networks.
\newblock \emph{IEEE/ACM Transactions on Computational Biology and Bioinformatics}, 17:\penalty0 56--, 01 2020.
\newblock \doi{10.1109/TCBB.2018.2853728}.

\bibitem[Annadani et~al.(2021)Annadani, Rothfuss, Lacoste, Scherrer, Goyal, Bengio, and Bauer]{DBLP:journals/corr/abs-2106-07635}
Yashas Annadani, Jonas Rothfuss, Alexandre Lacoste, Nino Scherrer, Anirudh Goyal, Yoshua Bengio, and Stefan Bauer.
\newblock Variational causal networks: Approximate bayesian inference over causal structures.
\newblock \emph{CoRR}, abs/2106.07635, 2021.
\newblock URL \url{https://arxiv.org/abs/2106.07635}.

\bibitem[Atanackovic et~al.(2023)Atanackovic, Tong, Hartford, Lee, Wang, and Bengio]{atanackovic2023dyngfn}
Lazar Atanackovic, Alexander Tong, Jason Hartford, Leo~J. Lee, Bo~Wang, and Yoshua Bengio.
\newblock Dyngfn: Towards bayesian inference of gene regulatory networks with gflownets, 2023.

\bibitem[Bengio et~al.(2021)Bengio, Jain, Korablyov, Precup, and Bengio]{bengio2021flow}
Emmanuel Bengio, Moksh Jain, Maksym Korablyov, Doina Precup, and Yoshua Bengio.
\newblock Flow network based generative models for non-iterative diverse candidate generation.
\newblock \emph{arXiv preprint arXiv:2106.04399}, 2021.

\bibitem[Bengio et~al.(2023)Bengio, Lahlou, Deleu, Hu, Tiwari, and Bengio]{gfn}
Yoshua Bengio, Salem Lahlou, Tristan Deleu, Edward~J. Hu, Mo~Tiwari, and Emmanuel Bengio.
\newblock Gflownet foundations.
\newblock \emph{Journal of Machine Learning Research}, 24\penalty0 (210):\penalty0 1--55, 2023.
\newblock URL \url{http://jmlr.org/papers/v24/22-0364.html}.

\bibitem[Bergen et~al.(2020)Bergen, Lange, Peidli, Wolf, and Theis]{velocitymethod}
Volker Bergen, Marius Lange, Stefan Peidli, F.~Wolf, and Fabian Theis.
\newblock Generalizing rna velocity to transient cell states through dynamical modeling.
\newblock \emph{Nature Biotechnology}, 38:\penalty0 1--7, 12 2020.
\newblock \doi{10.1038/s41587-020-0591-3}.

\bibitem[Bucur et~al.(2019)Bucur, Claassen, and Heskes]{B}
Ioan~Gabriel Bucur, Tom Claassen, and Tom Heskes.
\newblock Large-scale local causal inference of gene regulatory relationships.
\newblock \emph{International Journal of Approximate Reasoning}, 115:\penalty0 50--68, dec 2019.
\newblock \doi{10.1016/j.ijar.2019.08.012}.
\newblock URL \url{https://doi.org/10.1016%2Fj.ijar.2019.08.012}.

\bibitem[Chen and Liu(2022)]{causal_in_grn}
Guangyi Chen and Zhi-Ping Liu.
\newblock Inferring causal gene regulatory network via greynet: From dynamic grey association to causation.
\newblock \emph{Frontiers in Bioengineering and Biotechnology}, 10, 2022.
\newblock ISSN 2296-4185.
\newblock \doi{10.3389/fbioe.2022.954610}.
\newblock URL \url{https://www.frontiersin.org/articles/10.3389/fbioe.2022.954610}.

\bibitem[Chen et~al.(2018)Chen, Rubanova, Bettencourt, and Duvenaud]{od}
Tian~Qi Chen, Yulia Rubanova, Jesse Bettencourt, and David Duvenaud.
\newblock Neural ordinary differential equations.
\newblock \emph{CoRR}, abs/1806.07366, 2018.
\newblock URL \url{http://arxiv.org/abs/1806.07366}.

\bibitem[Chevalley et~al.(2023)Chevalley, Roohani, Mehrjou, Leskovec, and Schwab]{chevalley2023causalbench}
Mathieu Chevalley, Yusuf Roohani, Arash Mehrjou, Jure Leskovec, and Patrick Schwab.
\newblock Causalbench: A large-scale benchmark for network inference from single-cell perturbation data, 2023.

\bibitem[Chu et~al.(2020)Chu, Wang, Ma, Jia, Zhou, and Yang]{9338381}
Y.~Chu, X.~Wang, J.~Ma, K.~Jia, J.~Zhou, and H.~Yang.
\newblock Inductive granger causal modeling for multivariate time series.
\newblock In \emph{2020 IEEE International Conference on Data Mining (ICDM)}, pages 972--977, Los Alamitos, CA, USA, nov 2020. IEEE Computer Society.
\newblock \doi{10.1109/ICDM50108.2020.00111}.
\newblock URL \url{https://doi.ieeecomputersociety.org/10.1109/ICDM50108.2020.00111}.

\bibitem[Claassen and Heskes(2011)]{10.5555/3020548.3020565}
Tom Claassen and Tom Heskes.
\newblock A logical characterization of constraint-based causal discovery.
\newblock In \emph{Proceedings of the Twenty-Seventh Conference on Uncertainty in Artificial Intelligence}, UAI'11, page 135–144, Arlington, Virginia, USA, 2011. AUAI Press.
\newblock ISBN 9780974903972.

\bibitem[Cundy et~al.(2021)Cundy, Grover, and Ermon]{bcd}
Chris Cundy, Aditya Grover, and Stefano Ermon.
\newblock {BCD} nets: Scalable variational approaches for bayesian causal discovery.
\newblock \emph{CoRR}, abs/2112.02761, 2021.
\newblock URL \url{https://arxiv.org/abs/2112.02761}.

\bibitem[Dehghannasiri et~al.(2015)Dehghannasiri, Yoon, and Dougherty]{uncertainty}
Roozbeh Dehghannasiri, Byung-Jun Yoon, and Edward Dougherty.
\newblock Efficient experimental design for uncertainty reduction in gene regulatory networks.
\newblock 03 2015.
\newblock \doi{10.13140/RG.2.1.3252.2085}.

\bibitem[Deleu et~al.(2022)Deleu, G{\'o}is, Emezue, Rankawat, Lacoste-Julien, Bauer, and Bengio]{deleu2022bayesian}
Tristan Deleu, Ant{\'o}nio G{\'o}is, Chris Emezue, Mansi Rankawat, Simon Lacoste-Julien, Stefan Bauer, and Yoshua Bengio.
\newblock Bayesian structure learning with generative flow networks.
\newblock In \emph{Uncertainty in Artificial Intelligence}, pages 518--528. PMLR, 2022.

\bibitem[Denic et~al.(2009)Denic, Vasic, Charalambous, and Palanivelu]{uncertainty1}
Stojan Denic, B.~Vasic, Charalambos Charalambous, and Ravishankar Palanivelu.
\newblock Robust control of uncertain context-sensitive probabilistic boolean networks.
\newblock \emph{Systems Biology, IET}, 3:\penalty0 279 -- 295, 08 2009.
\newblock \doi{10.1049/iet-syb.2008.0121}.

\bibitem[Emmert-Streib et~al.(2014)Emmert-Streib, Dehmer, and Haibe-Kains]{grns}
Frank Emmert-Streib, Matthias Dehmer, and Benjamin Haibe-Kains.
\newblock Gene regulatory networks and their applications: understanding biological and medical problems in terms of networks.
\newblock \emph{Frontiers in Cell and Developmental Biology}, 2, 2014.
\newblock ISSN 2296-634X.
\newblock \doi{10.3389/fcell.2014.00038}.
\newblock URL \url{https://www.frontiersin.org/articles/10.3389/fcell.2014.00038}.

\bibitem[Glymour et~al.(2019)Glymour, Zhang, and Spirtes]{article_2}
Clark Glymour, Kun Zhang, and Peter Spirtes.
\newblock Review of causal discovery methods based on graphical models.
\newblock \emph{Frontiers in Genetics}, 10, 2019.
\newblock ISSN 1664-8021.
\newblock \doi{10.3389/fgene.2019.00524}.
\newblock URL \url{https://www.frontiersin.org/articles/10.3389/fgene.2019.00524}.

\bibitem[Huang et~al.(2019)Huang, Zhang, Zhang, Ramsey, Sanchez{-}Romero, Glymour, and Sch{\"{o}}lkopf]{cd}
Biwei Huang, Kun Zhang, Jiji Zhang, Joseph~D. Ramsey, Ruben Sanchez{-}Romero, Clark Glymour, and Bernhard Sch{\"{o}}lkopf.
\newblock Causal discovery from heterogeneous/nonstationary data.
\newblock \emph{CoRR}, abs/1903.01672, 2019.
\newblock URL \url{http://arxiv.org/abs/1903.01672}.

\bibitem[Karlebach and Shamir(2008)]{grn1}
Guy Karlebach and Ron Shamir.
\newblock Modelling and analysis of gene regulatory networks.
\newblock \emph{Nature reviews. Molecular cell biology}, 9:\penalty0 770--80, 10 2008.
\newblock \doi{10.1038/nrm2503}.

\bibitem[Lecca(2021)]{10.3389/fbinf.2021.746712}
Paola Lecca.
\newblock Machine learning for causal inference in biological networks: Perspectives of this challenge.
\newblock \emph{Frontiers in Bioinformatics}, 1, 2021.
\newblock ISSN 2673-7647.
\newblock \doi{10.3389/fbinf.2021.746712}.
\newblock URL \url{https://www.frontiersin.org/articles/10.3389/fbinf.2021.746712}.

\bibitem[Li et~al.(2023)Li, Xia, Chen, Zhao, Tao, and Chen]{10.1093/bib/bbad281}
Lin Li, Rui Xia, Wei Chen, Qi~Zhao, Peng Tao, and Luonan Chen.
\newblock {Single-cell causal network inferred by cross-mapping entropy}.
\newblock \emph{Briefings in Bioinformatics}, page bbad281, 08 2023.
\newblock ISSN 1477-4054.
\newblock \doi{10.1093/bib/bbad281}.
\newblock URL \url{https://doi.org/10.1093/bib/bbad281}.

\bibitem[Lorch et~al.(2021)Lorch, Rothfuss, Sch{\"o}lkopf, and Krause]{lorch2021dibs}
Lars Lorch, Jonas Rothfuss, Bernhard Sch{\"o}lkopf, and Andreas Krause.
\newblock Dibs: Differentiable bayesian structure learning.
\newblock \emph{Advances in Neural Information Processing Systems}, 34, 2021.

\bibitem[Madan et~al.(2022)Madan, Rector-Brooks, Korablyov, Bengio, Jain, Nica, Bosc, Bengio, and Malkin]{Madan2022LearningGF}
Kanika Madan, Jarrid Rector-Brooks, Maksym Korablyov, Emmanuel Bengio, Moksh Jain, Andrei~Cristian Nica, Tom Bosc, Yoshua Bengio, and Nikolay Malkin.
\newblock Learning gflownets from partial episodes for improved convergence and stability.
\newblock \emph{ArXiv}, abs/2209.12782, 2022.
\newblock URL \url{https://api.semanticscholar.org/CorpusID:252531657}.

\bibitem[Malkin et~al.(2022)Malkin, Jain, Bengio, Sun, and Bengio]{malkin2022trajectory}
Nikolay Malkin, Moksh Jain, Emmanuel Bengio, Chen Sun, and Yoshua Bengio.
\newblock Trajectory balance: Improved credit assignment in gflownets, 2022.

\bibitem[Mitrophanov and Groisman(2008)]{feedback_loop}
Alexander Mitrophanov and Eduardo Groisman.
\newblock Positive feedback in cellular control systems.
\newblock \emph{BioEssays : news and reviews in molecular, cellular and developmental biology}, 30:\penalty0 542--55, 06 2008.
\newblock \doi{10.1002/bies.20769}.

\bibitem[Murphy(2001)]{article_1}
Kevin Murphy.
\newblock Active learning of causal bayes net structure.
\newblock 06 2001.

\bibitem[Nguyen et~al.(2023)Nguyen, Mansouri, Madan, Nguyen, Ahuja, Liu, and Bengio]{nguyen2023reusable}
Trang Nguyen, Amin Mansouri, Kanika Madan, Khuong Nguyen, Kartik Ahuja, Dianbo Liu, and Yoshua Bengio.
\newblock Reusable slotwise mechanisms, 2023.

\bibitem[Pamfil et~al.(2020)Pamfil, Sriwattanaworachai, Desai, Pilgerstorfer, Beaumont, Georgatzis, and Aragam]{pamfil2020dynotears}
Roxana Pamfil, Nisara Sriwattanaworachai, Shaan Desai, Philip Pilgerstorfer, Paul Beaumont, Konstantinos Georgatzis, and Bryon Aragam.
\newblock Dynotears: Structure learning from time-series data, 2020.

\bibitem[Riba et~al.(2022)Riba, Oravecz, Durik, Jiménez, Alunni, Cerciat, Jung, Keime, Keyes, and Molina]{dataset}
Andrea Riba, Attila Oravecz, Matej Durik, Sara Jiménez, Violaine Alunni, Marie Cerciat, Matthieu Jung, Céline Keime, William Keyes, and Nacho Molina.
\newblock Cell cycle gene regulation dynamics revealed by rna velocity and deep-learning.
\newblock \emph{Nature Communications}, 13:\penalty0 2865, 05 2022.
\newblock \doi{10.1038/s41467-022-30545-8}.

\bibitem[Roy et~al.(2013)Roy, Das, Choudhury, Gohain, Sharma, and Bhattacharyya]{causal_in_grn2}
Swarup Roy, Dipankar Das, Dhrubajyoti Choudhury, Gunenja~G. Gohain, Ramesh Sharma, and Dhruba~K. Bhattacharyya.
\newblock Causality inference techniques for in-silico gene regulatory network.
\newblock In Rajendra Prasath and T.~Kathirvalavakumar, editors, \emph{Mining Intelligence and Knowledge Exploration}, pages 432--443, Cham, 2013. Springer International Publishing.
\newblock ISBN 978-3-319-03844-5.

\end{thebibliography}
\bibliographystyle{plainnat}

\newpage
\appendix
\section{Related Work}
\textbf{Bayesian Structure Learning} \quad  Recent advancements in differentiable Bayesian methods for static structure learning, including DiBS \cite{lorch2021dibs}, BCD-Nets \cite{bcd}, and DAG-GFlowNet \cite{deleu2022bayesian}, offer diverse graph parameterization approaches. While they excel in modeling uncertainty and structural distributions in smaller graphs, challenges arise when assuming natural dynamical systems adhere to  DAGs, particularly when cyclic structures from feedback mechanisms complicate the search space.

\textbf{Dynamic and Cyclic Structure Learning} \quad  Dynamic and cyclic structure learning has seen limited development. Before DynGFN \cite{atanackovic2023dyngfn}, the closest existing work in this area is CD-NOD \cite{cd}, which has shown potential for extension to handle cyclic graphs besides the original purpose of harnessing non-stationary data to unveil causal relationships in scenarios with changing generative processes over time. In contrast, traditional approaches to capturing intricate and uncertainty, including NeuralODEs \cite{od} that propose a sole explanatory structure, and DYNOTEARS \cite{pamfil2020dynotears} is a score-based approach for learning structure from time-series data. 

\textbf{Causal inference in large-scale biological systems}\quad  Among attempts to inferencing causal relations in large-scale biological systems, 
various methodologies have been developed in causal inference within large-scale biological systems. These approaches can be broadly categorized into three main groups: constraint-based methods \cite{9338381, 10.5555/3020548.3020565}, score-based methods \cite{chevalley2023causalbench, 10.3389/fbinf.2021.746712}, and hybrid methods \cite{B}. Constraint-based methods discern causal relationships by detecting statistical dependencies and independencies within the data. In contrast, score-based methods assign a score to each conceivable causal structure and select the structure with the most favorable score. On the other hand, hybrid methods amalgamate elements from constraint-based and score-based methodologies.

Recent years have witnessed a burgeoning interest in applying these causal inference techniques to extensive biological datasets \cite{chevalley2023causalbench}. For instance, researchers have harnessed these methods to unearth gene regulatory networks from gene expression data \cite{dataset}. These networks offer valuable insights into the regulatory mechanisms steering gene expression and can pinpoint potential targets for therapeutic interventions. Despite these noteworthy advancements, several formidable challenges persist in this domain. Among the principal challenges is grappling with confounding variables that can introduce spurious causal connections. Additionally, addressing missing data, a common issue in biological datasets remains a substantial hurdle to overcome.

\section{Bayesian dynamic structure learning framework}
\begin{equation} \label{ap:pgd}
   p(G, \theta,  \mathcal{D}) = p( \mathcal{D}|G,  \theta )p( \theta |G)p(G) 
\end{equation}
We adopt the framework from \citealp{atanackovic2023dyngfn} that decomposes the generative model as in Equation \ref{ap:pgd}. Subsequently, we employ the GFlowNet architecture to learn $P(G)$. Finally, we utilize the linear differential form $\frac{dx}{dt}=\mathbf{A}x$ to approximate the optimal $\theta$ to parameterize $P(\theta | G)$. 

\begin{figure}[b]
  \centering
  \includegraphics[width=0.9\textwidth]{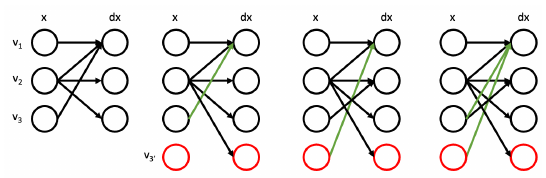}  
  \caption{\textbf{The model of indeterminacy.} A new variable that mirrors the values of $v_3$ is added, introducing three potential explanations for the data (in green). The visualization is adopted from DynGFN paper  \citep{atanackovic2023dyngfn}.}
  \label{fig:determinacy}
\end{figure}

\section{The Model of Indeterminacy} \label{ap:indeterminacy_model}
In this section, we summarize the model of indeterminacy, a strategy to generate the synthetic dataset investigated in this work, proposed by \citealp{atanackovic2023dyngfn} and visualized in Figure \ref{fig:determinacy}.

The purpose of the model of indeterminacy is to formulate a structure learning problem that contains many equivalent causal structures. 
Given the context of a dataset comprising pairs $(x, dx) \in \mathbb{R}^d \times \mathbb{R}^d$, which contains $d$ variables, a new variable is introduced to have $d+1$ variables in total. Specifically, the new variable replicates or is highly correlated with an existing variable $v$ and inherits the same parents as $v$. 
This extension results in the emergence of various potential explanatory graphs.
Finally, a sparsity penalty is applied to constrain the number of edges consistent in a valid graph and the number of possible cases. 

\section{Extended of Potential Future Work} \label{ap:extended}
\textbf{Decomposition} \quad  The current approach relies heavily on a monolithic design, which may not always be conducive to achieving a high level of generalization. In future work, we plan to develop a strategy that balances decomposition and parameter management. For instance, we aim to incorporate reusable mechanisms proposed in RSM \citep{nguyen2023reusable} to enhance the overall design.

\textbf{Design variants} \quad As illustrated in Figure \ref{fig:scalability}, although our proposed method exhibits superior scalability compared to other baselines, the performance of all models only marginally surpasses the random prediction threshold, as measured by the AUC metric. Therefore, further investigations and improvements in prediction quality are necessary. In addition to considering decomposition in the design, we will explore other avenues, such as enhancing sampling methods and the associated probability calculations, experimenting with alternative backbone models for GFlowNet beyond MLP, and exploring the potential benefits of fine-tuning.

\begin{algorithm}
\caption{Batch update training in \mname}\label{algo:main}
\begin{algorithmic}[1]
\State \textbf{Input}: Data batch ($x_b$, $dx_b$) 
\State $B$: batch of graphs computed parallelly
\State $n$: number of  \textbf{variables}  
\\

\State \textbf{GFlowNet Architecture}
\State $h_{id}$, $h_g$: encoded sizes of node's index and the whole graph, respectively
\State $h$: hidden size
\State $h = h_{id} + h_{g}$
\State $\mathbf{MLP_{id}} :  \mathbb{R}^{n} \rightarrow \mathbb{R}^{h_{id}}$
\State $\mathbf{MLP_{g}} :  \mathbb{R}^{n \times n} \rightarrow \mathbb{R}^{h_{g}}$
\State $\mathbf{MLP_{FW}} :  \mathbb{R}^{h} \rightarrow \mathbb{R}^{n+1}$ \Comment{n dimensions for forward probability, and the last dimension for state flow}
\State

\State \textbf{Model's computation flow}
\State \textbf{Inputs}: $node\_id \in \mathbb{R}^{B \times n}$, $graphs \in \mathbb{R}^{B \times n \times n}$
\State $rep_{id} = \mathbf{MLP_{id}}(node\_id)$
\State $rep_{g} = \mathbf{MLP_{g}}(graphs)$
\State $rep = \mathrm{cat}(rep_{id},\ rep_{g},\ dim=-1)$
\State $pred = \mathbf{MLP_{FW}}(rep)$
\State $log\_forward = pred[:, :-1].\mathrm{log\_softmax}()$
\State $log\_flow = pred[:, -1:]$ 
\State $log\_backward = \mathbf{0}_{B \times n}.\mathrm{log\_softmax}()$  \Comment{Uniform backward} 
\State \textbf{Outputs}: $log\_forward \in \mathbb{R}^{B \times n}$, $log\_backward \in \mathbb{R}^{B \times n}$, $flow \in \mathbb{R}^{B \times 1}$
\State 
\State \textbf{Step 0. Initialization}
\State $graphs \gets \mathbf{0}_{B \times  n\times n} $ \Comment{Empty $B$ graphs}
\State $node\_id \gets    \mathbf{0}_{B\times n}$  \Comment{Empty id of node of interest}
\State $done\_mask \gets    \mathbf{0}_{B\times n}$  \Comment{Empty visited mask}
\State $ll\_diff \gets    \mathbf{0}_{ n \times B }$ \\
\State \textbf{Training pipeline}
\For {$i$ in $0 \dots n$}  \Comment{ $ n + 1$ times of execution }

\State \textbf{Step 1. GFlowNet computation}
\State $log\_forward,\ log\_backward,\ log\_flow = \mathbf{model}(node\_id, graphs) $ 
\State
\If {i > 0}  \Comment{Excluding the starting node} 
\State \textbf{Step 2. Sampling \textit{all} incoming edges}
\State $actions \gets \mathrm{sample\_all\_incomming\_edges(log\_forward)} $
\State $graphs[:, node\_id] \gets actions $
\EndIf
\State 
\State \textbf{Step 3. Updating flows} 
\State  $ll\_diff[i]\ += log\_flow$
\State $ll\_diff[i]\ += log\_forward.\mathrm{gather}(actions)$
\If {i > 0}
\State $ll\_diff[i-1]\ -= log\_flow$
\State $ll\_diff[i-1]\ -= log\_backward.\mathrm{gather}(actions') $
\State $actions' \gets actions$ \Comment{Preparing for updating flows in the next turn}
\EndIf
\If {$i$ ==  $n$} \Comment{Reaching the last turn}
\State $log\_rewards \gets -||dx_b - \widehat{dx_b}||^2_2 + \lambda_0||graphs||_0 $
\State $ll\_diff[node\_id]\ -= log\_rewards $
\Else 
\State \textbf{Step 4. Sampling the next node of interest} 
\State $node\_id = (log\_forward - done\_mask \times \mathrm{inf}).\mathrm{argmax()}$ \Comment{Binary matrix}
\State $done\_mask += node\_id$ 
\EndIf
\State

\EndFor
\State

\State \textbf{Optimization }
\State $\mathcal{L}_{DB}  = ll\_diff^2.\mathrm{mean()}$


\end{algorithmic}
\end{algorithm}

\end{document}